# Towards automated high-throughput screening of *C. elegans* on agar

Mayank Kabra[1], Annie L. Conery[2], Eyleen J. O'Rourke[2,3], Xin Xie[4], Vebjorn Ljosa[5], Thouis R. Jones[5], Frederick M. Ausubel[2,3], Gary Ruvkun[2,3], Anne E. Carpenter[5], and Yoav Freund[4,*]

[1]ECE, University of California, San Diego

[2]Department of Molecular Biology and Center for Computational and Integrative Biology, Massachusetts General Hospital

[3]Department of Genetics, Harvard Medical School

[4]CSE, University of California, San Diego

[5]Imaging Platform, Broad Institute of Harvard and MIT

**ABSTRACT**

High-throughput screening (HTS) using model organisms is a promising method to identify a small number of genes or drugs potentially relevant to human biology or disease. In HTS experiments, robots and computers do a significant portion of the experimental work. However, one remaining major bottleneck is the manual analysis of experimental results, which is commonly in the form of microscopy images. This manual inspection is labor intensive, slow and subjective. Here we report our progress towards applying computer vision and machine learning methods to analyze HTS experiments that use *Caenorhabditis elegans* (*C. elegans*) worms grown on agar. Our main contribution is a robust segmentation algorithm for separating the worms from the background using brightfield images. We also show that by combining the output of this segmentation algorithm with an algorithm to detect the fluorescent dye, Nile Red, we can reliably distinguish different fluorescence-based phenotypes even though the visual differences are subtle. The accuracy of our method is similar to that of expert human analysts. This new capability is a significant step towards fully automated HTS experiments using *C. elegans*.

## 1 INTRODUCTION

### 1.1 High-throughput screening

An emerging method of studying complex genetic pathways and discovering drugs and their targets is to use model organisms for high-throughput screening (HTS). HTS uses automation to allow labs to conduct a large number of similar tests in an efficient, fast and cost-effective manner. In HTS for genomic analyses, a separate gene inactivation is conducted for every single gene in the model organism. Each gene is silenced using RNA interference (RNAi) to test whether it affects a given genetic pathway. In HTS experiments for drug discovery individual compounds from a large library of chemicals are tested. The number of genes or compounds tested in single screen starts at about 10,000 and can go up to the millions. Individual gene inactivations or compound tests are performed in a few millimeters wide "well" of a "plate" containing 96 or 384 wells. The wells are filled by robots, incubated, and then imaged using an automated microscope. The recent availability of affordable HTS equipment has made conducting such high-throughput studies more common.

### 1.2 *C. elegans*

Currently, most HTS experiments are conducted using cells in culture; generally a single cell line. A single well in a screening plate will typically contain thousands of identical cells, producing highly consistent and reliable results. However, many important phenomena, such as aging and immunity, can only be comprehensively studied in multicellular organisms. *C. elegans* is a free-living, 1mm long, transparent nematode (roundworm), which is found in compost. Studies based on *C. elegans* have led to major discoveries, including the identification of key genes in the insulin pathway (Kenyon et al, 1993, Ogg et al, 1997), and the targets of common drugs, such as Prozac (Ranganathan et al, 2001). *C. elegans* also allowed the discovery of fundamental and conserved biological processes such as programmed cell death (Ellis et al, 1986, Nobel Prize 2002) and gene regulation by small RNAs (Fire et al, 1998, Wightman et al, 1993, Lee et al, 1993, Nobel Prize 2006). *C. elegans* is attractive for HTS not only for the conservation of genetic pathways relevant to human disease but because worms are small (many worms can fit in a single well), transparent (thus easy to image), and easy to grow and manipulate. Silencing individual genes is achieved by feeding the worms with bacteria that express double stranded RNA (dsRNA) targeting a given gene, making *C. elegans* the only animal in which full-genome screens are feasible. *C. elegans* has also shown promise in the search for new therapeutic compounds. Novel anti-infectives have been recently described that, unlike traditional antibiotics, do not directly kill pathogenic bacteria but function as curative agents by enhancing the immune system of the infected host (Moy et al, 2009).

* To whom correspondence should be addressed (yfreund@cse.ucsd.edu)

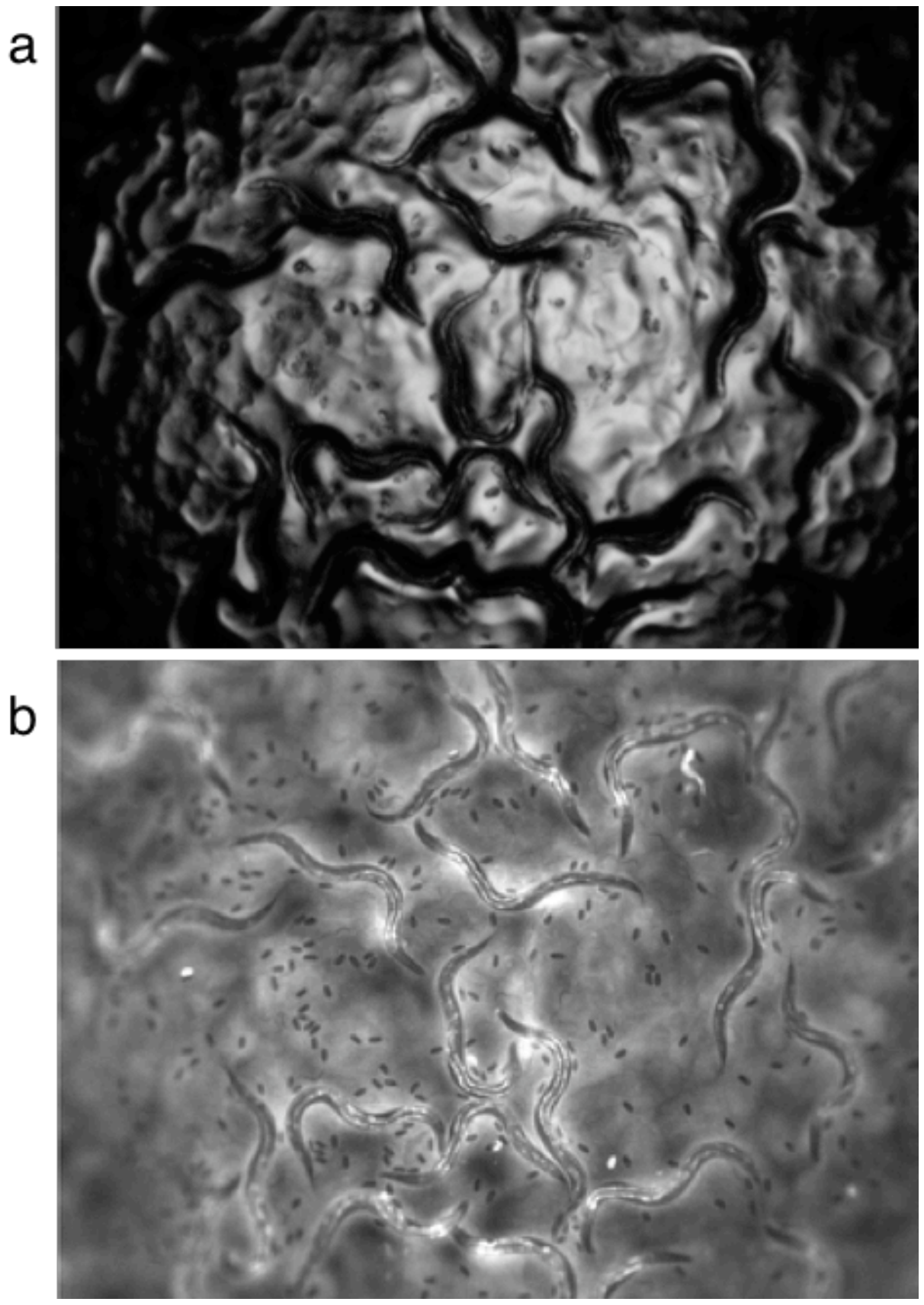

Figure 1: Examples of brightfield and fluorescent images. a) Brightfield image of the whole well. Limitations of the automated imaging setup cause the edges of the image to be out of focus and dim. The tracks left in the bacterial food source, and touching and overlapping worms complicate segmenting the worms. b) Fluorescent image of the same well. The vital dye Nile Red stains a lysosomal compartment along the worm's intestine. The phenotypes "high Nile Red", "wild type" or "low Nile Red" are distinguished based on the intensity and distribution of the fluorescent signal.

### 1.3 Planned HTS experiments

It has been recently proposed that the intensity of the fluorescent dye Nile Red (**NR**) correlates with the *C. elegans* rate of aging (O'Rourke et al, 2009). Worms with low Nile Red signal age more slowly, and animals with increased Nile Red signal live shorter than wild type animals. We are currently in the planning stage of HTS experiments on *C. elegans* based on Nile Red.

The HTS experiments are analyzed using two types of images: standard normal light or brightfield images (BF) and fluorescence (FL) images (Figure 1). BF images inform about developmental and gross anatomical defects. FL images exhibit the distribution of Nile Red dye in the bodies of the worms. Till now, these images were analyzed manually. Manual analysis is labor intensive, subjective, and error prone. Our goal is to minimize these problems by automating the analysis of worm images using computer vision algorithms. In these experiments the worms are grown on agar. The agar background significantly complicates segmentation i.e., separating the worms from the background. Segmentation is difficult because the worms appearance is similar to tracks that they leave in the agar (Figure 1a).

When the worms are imaged in liquid media, the segmentation is much simpler. However there are inherent problems with growing worms in liquid medium. Worms grown in liquid medium grow slowly and are stressed (Houthoofd et al, 2002). In addition, RNAi is less effective in liquid media (Lehner et al, 2006). In past experiments (Moy et al, 2009) the worms were grown in agar and then transferred to liquid media for imaging and analysis. However, transferring worms to liquid is labor-intensive. In addition, the stress of transfer can affect the worms and invalidate the experimental results. Our plan is to image the worms in the agar. This will greatly facilitate screening and additionally, enable screens that cannot have worms in liquid at any stage due to the stress.

### 1.4 Worm Segmentation

Our main contribution is a novel method for segmenting worms from background when the worms are in agar. Our approach for segmenting worms in agar is to combine low-level image processing and scoring functions generated by machine learning algorithms. The low-level image processing performs basic operations such as smoothing, edge detection, connected element analysis and feature calculation (area, average brightness etc.). The scoring functions takes as input the features calculated by the low-level image processing and outputs a score for each pixel. High scores correspond to pixels that are inside the worms and low scores correspond to background pixels. The scoring function is generated using machine learning algorithms

The reason we use machine learning is that transforming low-level image features into segmentation score is a very complex task. An experienced person can segment the worms in an instant. However, the mental process of segmentation is very hard to describe in words and therefore very hard to translate into computer code. The machine learning approach avoids the need to describe this mental process. Instead, we have the experienced person mark the boundaries of the worms in a small number of images. These annotated images are used as *training data*. The machine learning algorithm analyzes the training set and constructs a scoring function. A good scoring function should perform well both on the training data and on new images. We use the Adaboost learning algorithm (Freund et al, 1997), which has been used successfully in the past for object detection tasks (Viola et al, 2001; Liu et al, 2008).

To the best of our knowledge, ours is the first effective computer vision method for distinguishing worm bodies from background on the agar. It is a robust method that works effectively for a large variety of images produced in different experimental conditions.

### 1.5 Automated phenotyping

The aim of analyzing images in a HTS experiment is to identify the phenotype present at the end of the experiment. We developed an automatic method for classifying images according to the phenotype of the worms in the image. We classify Nile Red phenotypes into 3 classes: hNR (high-Nile Red), wild type, lNR (low-Nile Red). In order to identify the Nile Red phenotype of the worm, we combine the segmentation results with information from the fluorescent image.

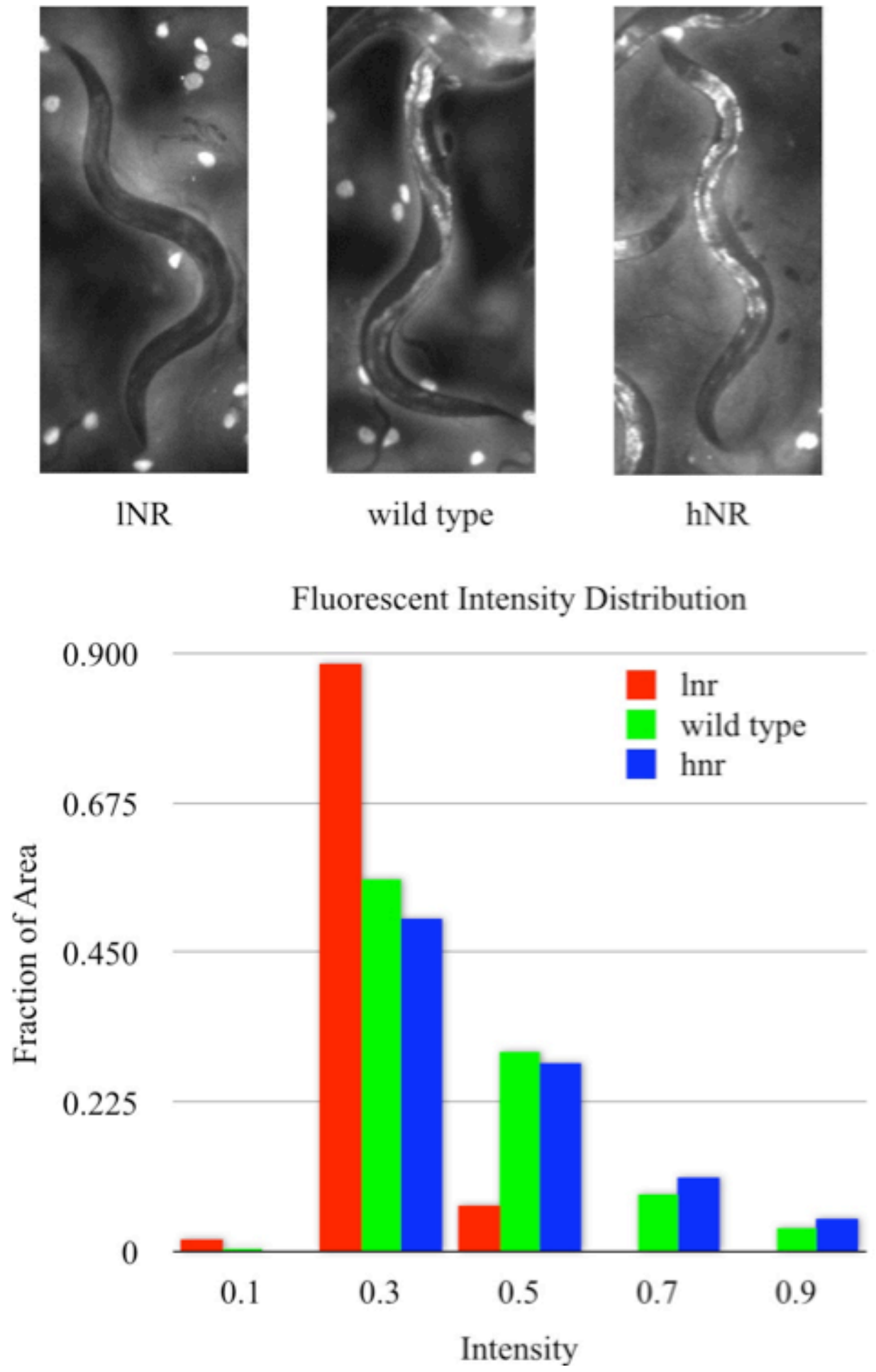

Figure 2: Nile Red fluorescence intensity measurements. Worm phenotypes are separated by evaluating the Nile Red fluorescent signal within the worms. The fluorescent intensity distribution inside the worms is similar, which suggests that the differences between phenotypes are subtle and cannot be captured by measuring the fluorescent intensity.

Automatic identifying of phenotypes is typically very simple when processing images of cell-cultures. The average intensity of the fluorescent dye within each cell provides the required information.

We tested whether the average intensity of the dye can be used to identify Nile Red phenotypes in *C. elegans*. We found out that the average intensity of Nile Red within the body of the worm has a very weak correlation with the known phenotype (Figure 2). On the other hand the phenotype can be reliably identified by human observation. Further study revealed that the key to identifying the phenotype is that Nile Red is distributed in two stripes in each worm, corresponding to regions of the worm's gut, where the lysosome-related organelles that contain the dye are concentrated. We found that it is relatively easy to segment the Nile Red stripes in the fluorescent image. By combining features of the stripes with features of the worm within which the stripe appears we were able to reliably identify the phenotype of the worm. In particular, we found that the ratio between the area of the stripe and the area of the surrounding worm is a reliable feature for identifying the worm's phenotype.

To quantify the predictive performance we used images from plates in which each well contained worms with a known phenotype (hNr, lNR or wild type). We evaluated the accuracy of our phenotype classifier using cross validation (CV). The error rate for discriminating lNR vs. wild type is about 1% and for hNR vs. wild type it is about 5%. Both accuracy levels are similar to the accuracy levels achieved by two biologists manually scoring the images.

The main concern in analyzing the results of screening is to ensure that no gene inactivation with a phenotype different from wild type is missed. Experiments (wells) that produce ambiguous results are usually repeated. Taking this into consideration, we conducted a second experiment in which we allowed the classification algorithm to abstain and not predict the phenotype in some of the wells. Using a classifier with abstention, we obtained the following results. For lNR vs. wild type the classifier abstained on 5% of the wells and made no mistakes in classifying the remaining 95% of the images. For hNR vs. wild type the classifier abstained on 25% of the wells and had a 1% error rate on the rest of the images. The error rate with the abstentions is significantly lower than with manual scoring. We conclude that the proposed method is sufficiently accurate for automating the analysis of the planned Nile Red HTS experiments.

### 1.6 Related Work

Previous work on *C. elegans* that analyzed adult worms was done on time-lapse movies of single worm (Feng et al, 2004, Cronin et al, 2004, Hoshi et al, 2006, Cronin et al, 2005) or multiple worms (Fontaine et al, 2006, 2007; Tsechpenakis et al, 2008; Romot et al, 2008; de Bono et al, 1998) in a Petri dish. There is also related work on adult *C. elegans* using high-resolution confocal microscopy images of individual worms. The worms are straightened (Peng et al, 2008) to create a digital atlas of individual cells of the *C. elegans* adult that is used to study cell lineage by segmenting nuclei (Liu et al, 2009). Algorithms also exist for analyzing three-dimensional, time-lapse movies of the individual nuclei in a *C. elegans* embryo (Bao et al, 2006).

The most similar work to date was the *C. elegans* high-throughput screen scored by automated image analysis for a drug screen looking for inhibitors of infection, where worms that were grown on agar were washed from their bacterial food source and transferred to liquid (Moy et al, 2009). The image analysis was done using CellProfiler (CellProfiler, 2009) using a complex image-processing pipeline.

## 2 METHODS

### 2.1 High level software design

The main technical challenge facing the automated analysis of images of *C. elegans* on agar is the complexity and variability of the images. Two causes of complexity in BF images (see Figure 1a) are the tracks left by the worms and the meniscus created when the agar is poured into the wells. The curved meniscus surface allows only the central part of image to be properly lit and in focus. The fluorescent images of the Nile Red dye (Figure 1b) are less complex than the BF images because the tracks are less visible and the meniscus only has the effect of putting the edge of the well out of focus and does not affect the fluorescence. On the other hand, the signal in the FL images varies greatly with experimental conditions and across different plates within the same experiment. Our approach segments the worms in the complex but less variable BF images and then uses this segmentation to identify the Nile Red phenotype from the FL images.

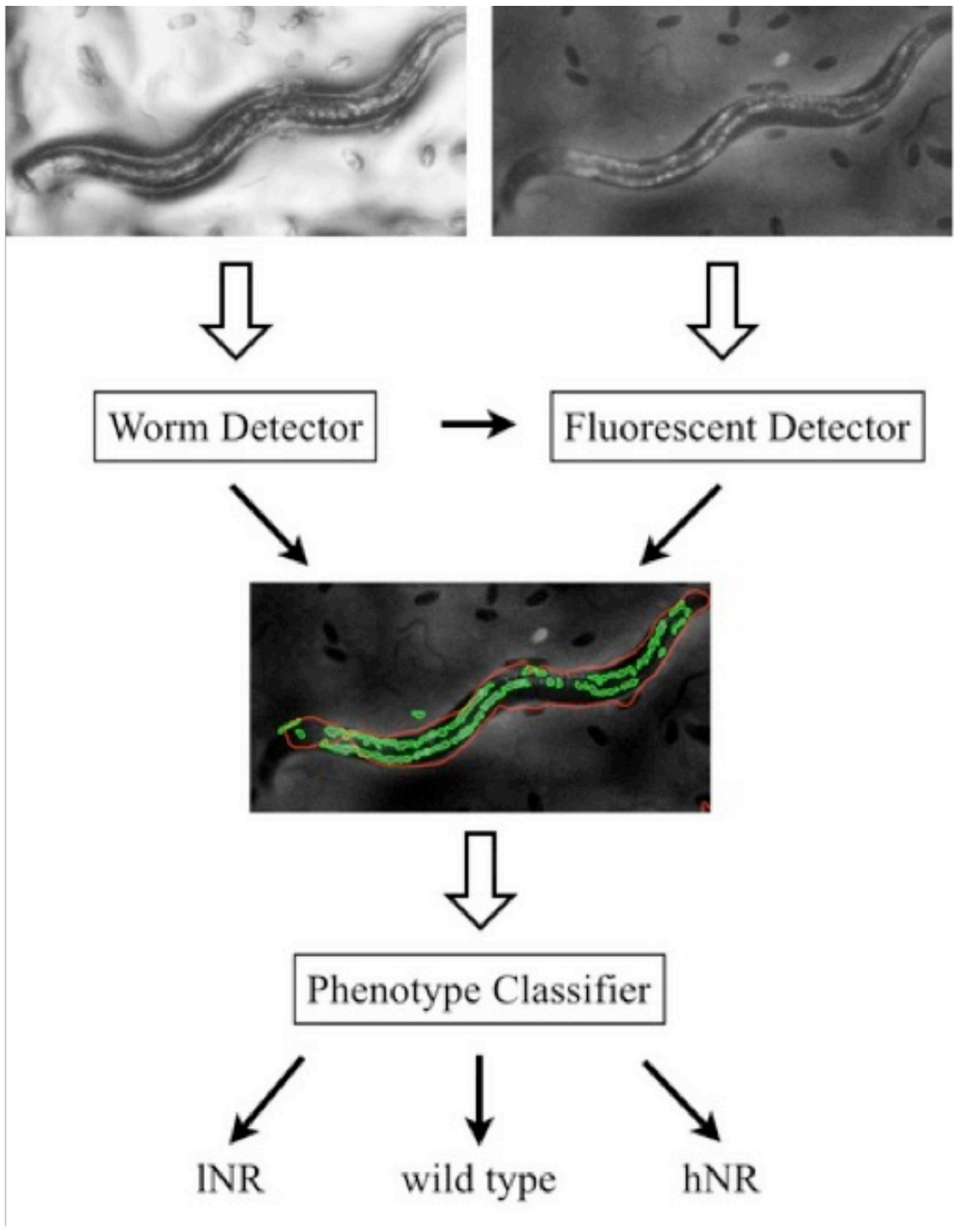

Figure 3: High-level design of the system. The worms are detected in the brightfield image using the worm detector. In the fluorescent image, we detect Nile Red stripes that mark the lysosome-related organelles along the worm's intestine. The outputs of these two detectors are then combined to classify worms into the three phenotypes: high Nile Red, wild type, and low Nile Red.

The high-level design of our system is illustrated in Figure 3. The system consists of three components: a worm detector, a fluorescence detector and a phenotype classifier. The worm detector segments the worms using the BF image. The fluorescence detector detects the areas of Nile Red concentration along the worm's intestines. Finally, the phenotype classifier classifies each well according to the Nile Red phenotype of the worms in that well: wild type, high Nile Red (hNR) and low Nile Red (lNR).

Table 1: Comparison of segmentation inaccuracies of techniques that use local information as implemented in CellProfiler and our worm detector. The inaccuracy is calculated as the mismatch between manually labeled worms and the segmentation results over 18 images. The mismatch is reported relative to the area of the ground truth. The worms were segmented in CellProfiler by thresholding after contrast adjustment.

| Segmentation Method | CellProfiler | Our worm detector |
|---|---|---|
| False Positive | 104% | 25% |
| False Negative | 35% | 30% |
| Total Mismatch | 139% | 56% |

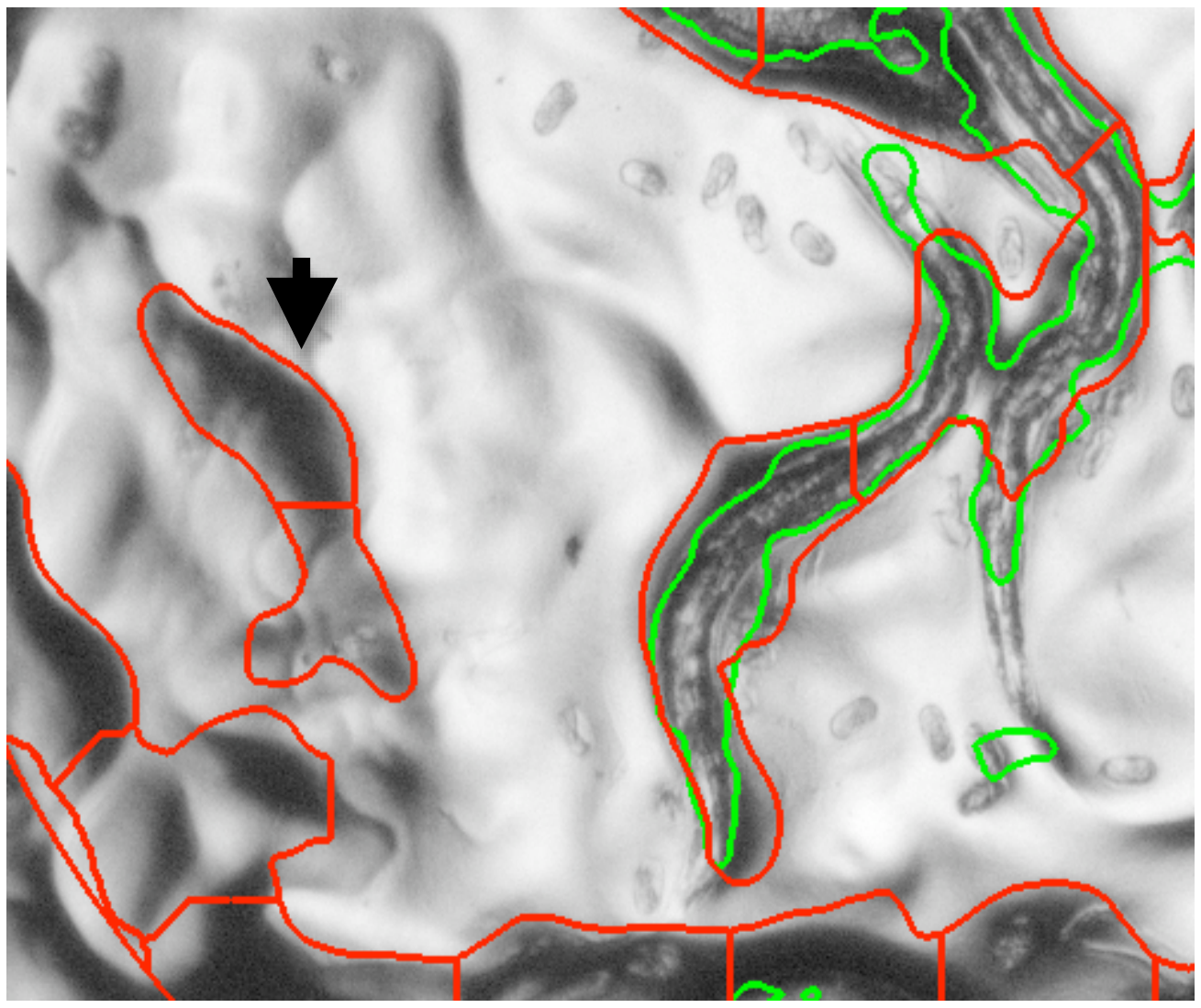

Figure 4: Comparison of segmentation results from two analysis systems. The red outlines are the segmentation results from CellProfiler and the green outlines are segmentation results from our segmentation method. In CellProfiler, the image is first normalized and then segmented by thresholding. Such segmentation methods that use only local information mark tracks (arrowhead) as worms. Our method uses the shape of the worm and visual cues such as texture, contrast and worm's edges. The segmentation results are overlaid over log-transformed image for better visualization.

Each of the three components utilizes machine learning. The main advantage of using machine learning over manual tuning of different detectors is that labeling examples does not require an understanding of the computer vision algorithms and can be integrated seamlessly into the existing experimental protocols. This is particularly important for the phenotype classifier. The visual differences between Nile Red phenotypes in the FL images are subtle and are easily masked by the differences between experimental conditions. Therefore, in order to accurately classify the phenotype we needed to calibrate the phenotype classifier separately for each plate. In an actual screening the calibration can be automated because usually a few of the wells in each plate are devoted to contain worms of a known phenotype (no gene inactivation). Using these calibration wells to train the phenotype classifier we can then reliably identify the phenotype of the other wells without any manual intervention.

In the next sections we describe the methods behind each of the three components.

### 2.2 Worm segmentation

As described earlier, segmentation of *C. elegans* worms in agar is a very challenging problem. The worms leave deep tracks in the agar and it is hard to distinguish the tracks from the worms. Another image analysis program, CellProfiler, whose segmentation methods perform very well for cells in culture and worms in liquid, encounters severe problems when used to segment worms in agar as can be seen in Figure 4 and Table 1. In order to segment the *C. elegans* worms reliably and accurately, a large number of visual cues must be combined. We do this by constructing a scoring function that identifies short segments of the worm. The segmentation of the worms is based on this scoring function. The scoring function itself is generated using a machine learning algorithm.

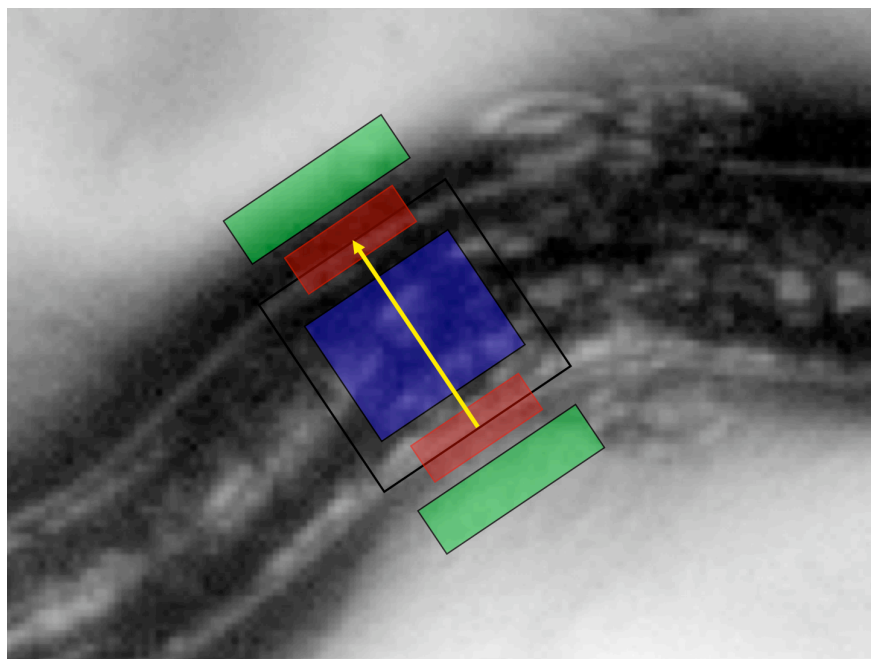

Figure 5: Example of worm segment. The features for detecting worm segments are built by measuring responses of filtered images inside the rectangles.

We represent the shape of a worm using a sequence of line segments that are orthogonal to the worm's midline and extend along the width of the worm (Figure 5). *Worm segments* are relatively large visual objects, measuring around 20 pixels in size, and contain more visual information than the commonly used pixel-based segmentation algorithms. We formulate the task of identifying worm segments as a machine learning problem by collecting a training set consisting of positive examples — line segments that extend across a worm and negative examples — line segments that do not satisfy this condition. Using the AdaBoost learning algorithm we generate a scoring function, which maps the image around the line segment to a real value. The value of the scoring function is to be a large positive number for line segments that cross a worm and small or negative otherwise. This property should hold for both examples in the training set (low training error), and, more importantly, it should hold for examples that are not in the training set but are generated from the same distribution (test error). If the difference between the training error and the test error is large, we say that the machine learning method over-fits. An important advantage of using Adaboost is that it tends not to over-fit (Schapire et al, 1998).

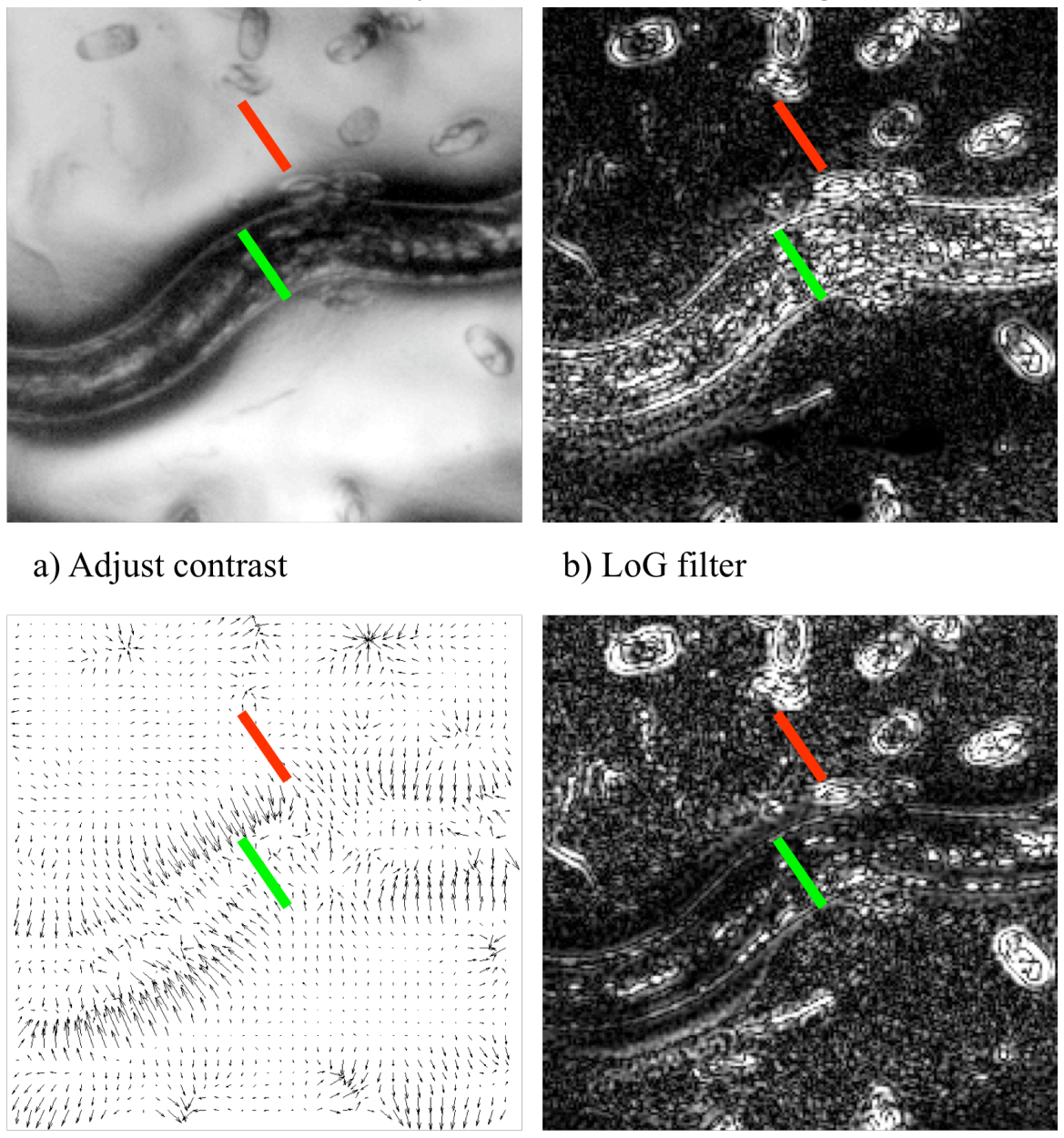

a) Adjust contrast b) LoG filter

c) Sobel filter d) Adjust contrast + LoG filter

Figure 6: Worm segment detection based on visual cues. These visual cues can be quantified using filtered responses around the segments. The filter responses are different around the green worm segment and the red segment on agar. By measuring these filtered responses around the segments we build features that are used by the Adaboost classifier to detect the segments.

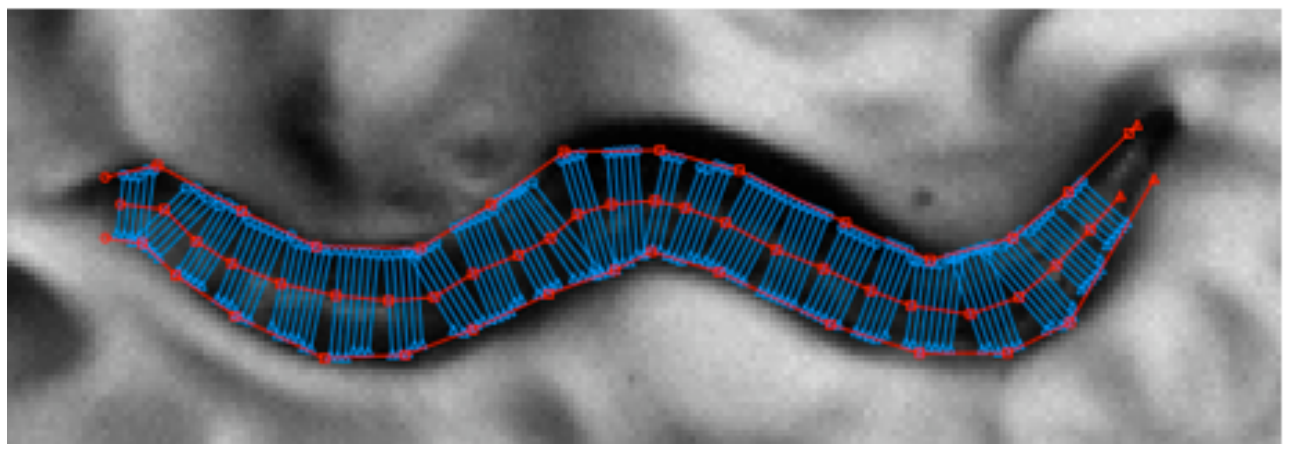

Figure 7: Worm segments. Worms are segmented in the brightfield image by detecting worm segments. The worm segments (blue lines) extend from one side of the worm to the other side and are orthogonal to the median line. For training the detector, the user outlines the worms (red). The worm segments are then automatically generated.

The low-level features are the outputs of linear filters including the Sobel edge detector and the Laplacian of Gaussian operator. Figure 6 depicts the outputs of some of these filters. The features that serve as inputs to the scoring function are quantiles of the filtered values in rectangular regions whose locations are defined relative to the location of the line segment that is to be scored (see Figure 5). We use quantiles rather than averages because of their superior statistical stability.

To train the scoring function, the user annotates a small number of worms (around 10) by drawing lines that mark the sides and the midline of the worms. Using these line annotations, we generate an initial training set. The positive examples are the line segments defined by the human annotation (Figure 7) and the negative examples are randomly selected line segments of the expected length (10-30 pixels). To generate the scoring function we use the Adaboost learning algorithm as implemented in the JBOOST open-source software package (Jboost). The initial training set does not generate a sufficiently accurate scoring function. The reason is that the initial negative examples, chosen uniformly at random, do not give sufficient representation to the "non-worm" segments that are difficult to classify. In order to overcome this problem, we add to the training set negative line segments whose score is high to the training set (Figure 8) and we also increase the number and variety of images in the training set. At the end of a long refinement process, which lasted the better part of a year, we created a scoring function whose scores are accurate, reliable, and effective for a large variety of BF images (Xie, 2009).

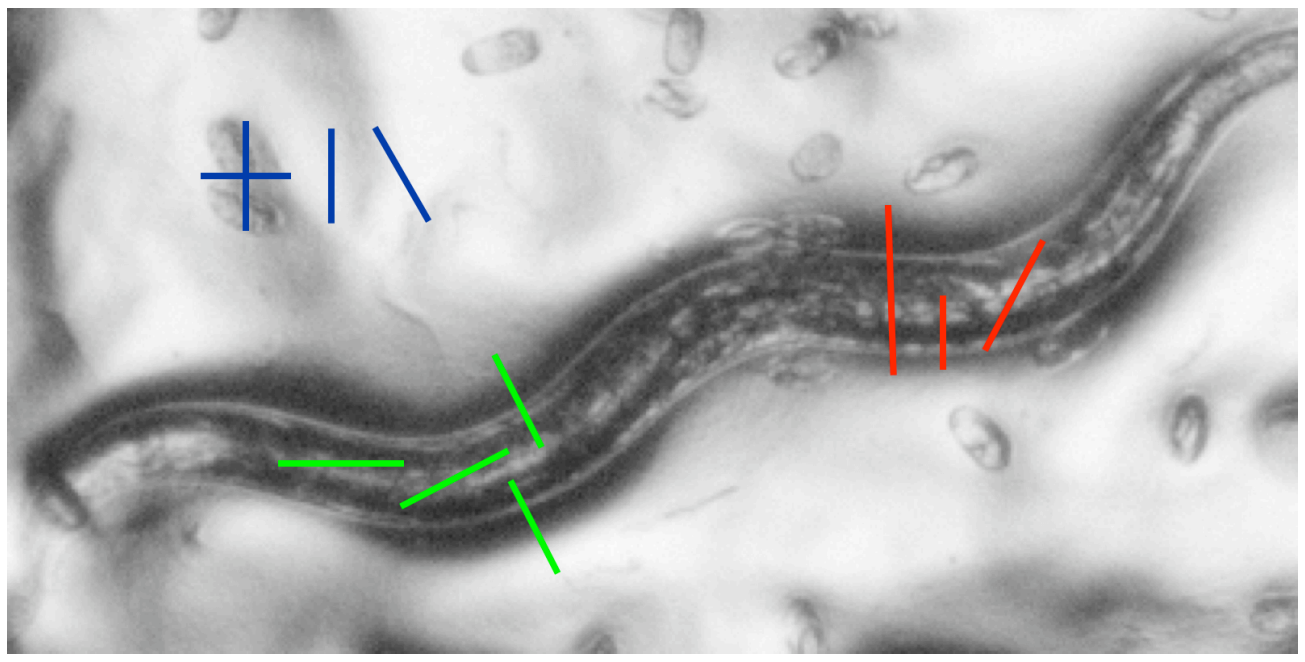

Figure 8: Negative worm segments examples. To improve worm detectors performance we progressively add difficult negative examples. We start by choosing random segments as negative examples. Most of the segments are similar to blue segments (easy). Later we add segments similar to green (moderately difficult) and red (very difficult).

Our segmentation method is based on this scoring function. Given an image of 1388x1036 pixels our segmentation algorithm calculates the score for line segments in 1388x1036 locations and at every $30^0$ angle. For each location we retain the highest score, corresponding to the most highly scored angle. The segmentation of the image into worms vs. background is done by thresholding these scores (Figure 9).

The computation required for this segmentation process is currently the main bottleneck of our processing pipeline. Segmenting a 1388x1036 pixels image requires about one hour of processing on a workstation. We are currently working on a hardware-based implementation of the segmentation algorithm. We expect this hardware implementation will speed up the computation by a factor of 100.

## 2.3 Fluorescence Detector

The average intensity of fluorescence inside cells has proven to be a very reliable signal for distinguishing phenotypes in HTS using cell cultures. Unfortunately, it appears that the average intensity of

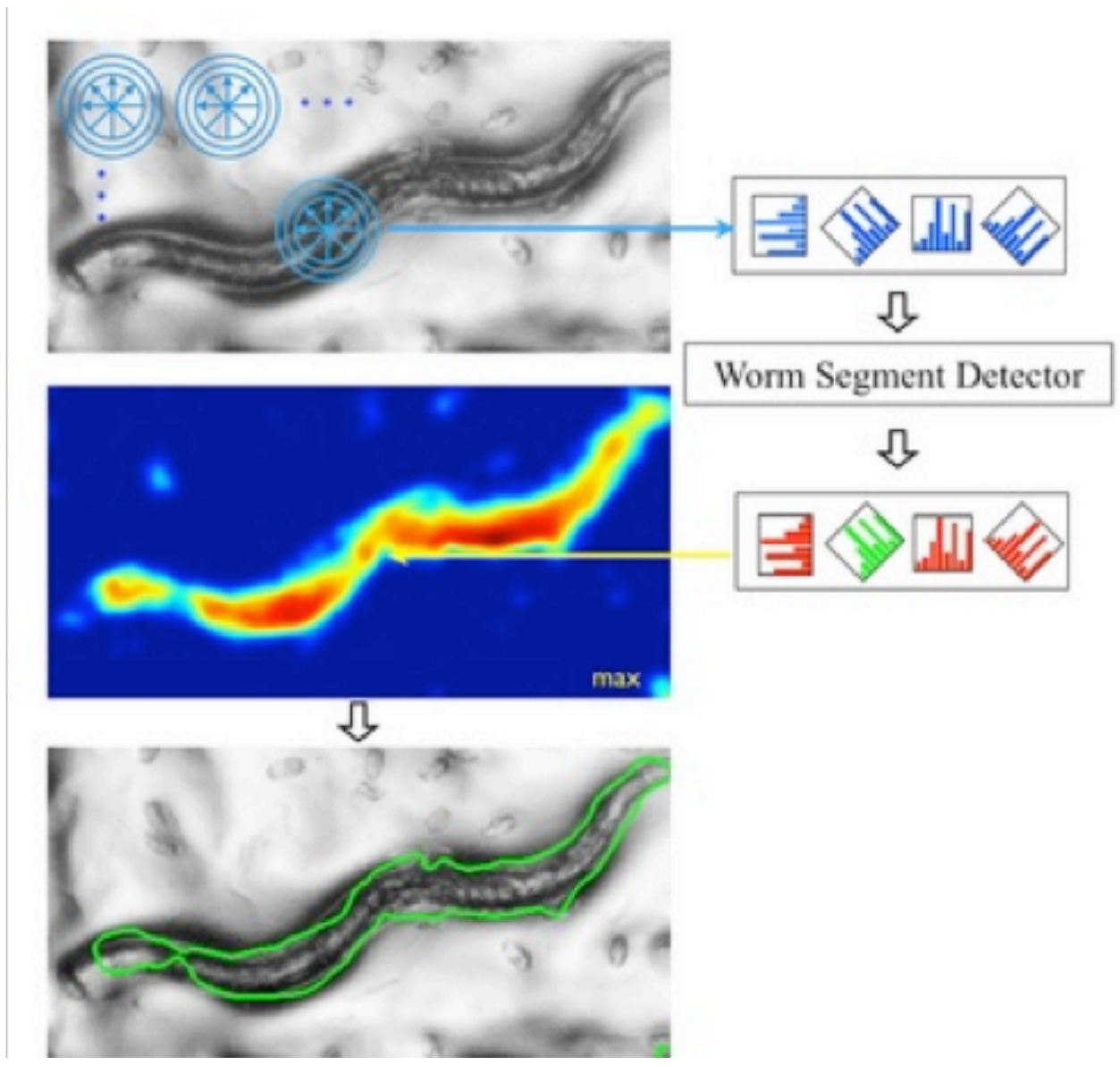

Figure 9: Worm segment detector. At each pixel location, we classify segments with different orientations and different sizes. The worm segment detector gives a score to each segment, where a positive score indicates that the segment is indeed a worm segment. For each pixel location, we find the segment that gets the highest score. These scores are color-mapped to generate the score image. The positive regions of the score image when outlined mark the boundaries of the worm.

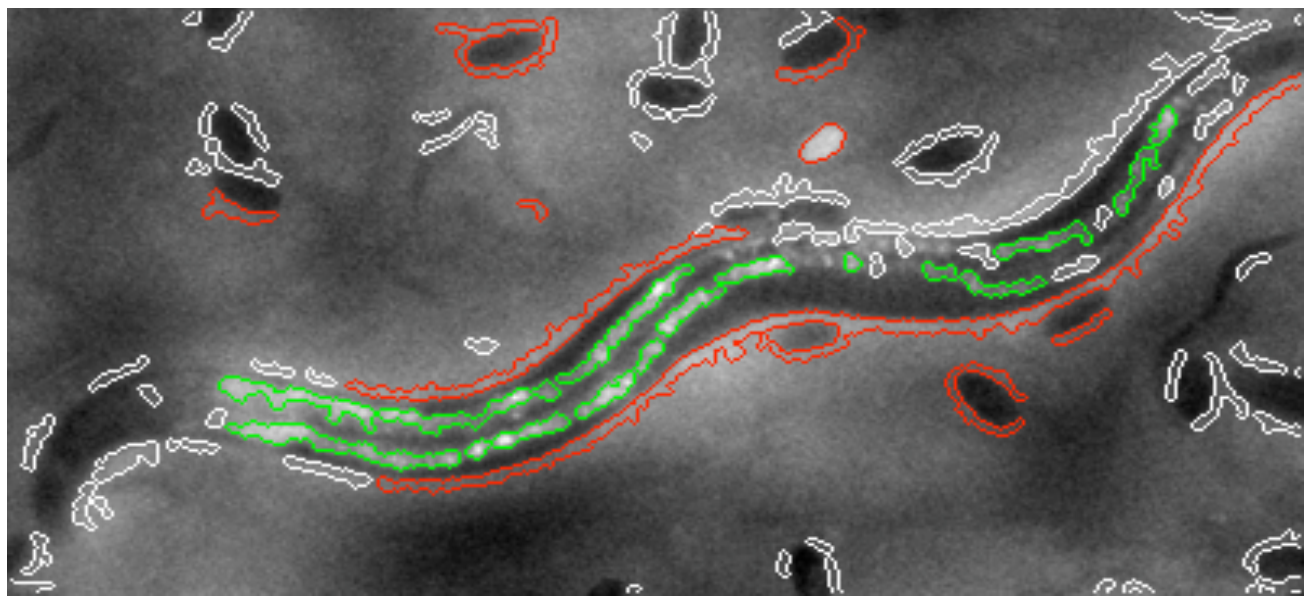

Figure 10: Nile Red marks the lysosome-related organelles along the worm's intestine and the staining pattern appears as stripes in the fluorescent image of the worm. The stripes can be outlined using a blob detector and separated using the fluorescence stripe detector.

Nile Red in the area of the worm is only weakly correlated to changes in the phenotype even when the worm segmentation is done manually (Figure 2). On the other hand, biologists can reliably identify the phenotypes of worms by inspecting the NR images. The reason that the average intensity is a poor predictor is that the Nile Red signal is not homogeneously distributed inside the worm, but is concentrated in lysosome-related organelles within the intestine (Schroder et al, 2007; O'Rourke et al, 2009).

The Nile Red signal appears as two bright stripes along the length of the worm (Figure 1b, 2). The appearance of the two stripes represents two parallel stripes of intestinal cells separated by the lumen. The discrimination of Nile Red phenotypes based on the appearance of these stripes is much more reliable. In lNR phenotype, the stripes are dim and more homogeneous as compared to wild type, while in the hNR phenotype the stripes are patchier, brighter and occupy a larger area (Figure 2).

To identify the stripes, we filter the image with Laplacian of Gaussian (LoG) filter (Marr et al, 1980) and threshold the filtered image. This operation correctly segments the stripes, but it also generates other blobs, corresponding to other bright regions in the fluorescent image. We separate the stripes from the other objects using a classifier generated by Adaboost (Figure 10). This classifier uses properties of the identified blobs, as well as their relationship to the worm segments detected using the BF image, an input.

The training set for stripe classifier is collected in an interactive manner. On the FL image, we show the outline of the segmented objects and then label the stripes. The features for classification are based on the geometry, shape and intensity profile of the object. Using these features we found that we had to label stripes in only 6 fluorescent images to generate a classifier with sufficient accuracy. The segmentation and detection of stripes is sufficiently stable to allow accurate analysis of images produced by experiments performed a year after the experiments used to train the stripe detector.

## 2.4 Phenotype Classifier

The appearance of the worms and the appearance of the fluorescent stripes are consistent and stable enough to allow reuse of the segmentation and detection algorithms. On the other hand, the appearance of different phenotypes is highly dependent on the experimental conditions and can change between different days because of slight variations in the experimental conditions between batches of plates prepared on different days. In order to achieve reliable classification of the phenotype, we found it necessary to retrain the phenotype classifier for each plate. However, manual scoring also requires control wells in each plate, consequently the requirements for automated analysis do not increase the experimental labor or cost. Additional experiments will be required to determine whether the same classifier can be used for different plates from the same batch. As with the other components, we use Adaboost to train the phenotype classifier.

Table 2: Comparison of the automated phenotype separation method with humans. Each set of images is divided into half, where set 1 is used for training and set 2 is used as test. Human experts repeated the experiments on 4 such sets. The automated method was evaluated on 20 such sets.

| Type | Plate number | Experimentalist 1 Error (%) | Experimentalist 2 Error (%) | Automated method's error - without abstention (%) | Automated method's error – with abstension (%) | Examples on which we abstain (%) |
|---|---|---|---|---|---|---|
| wild type vs lNR | 1 | 1.6 | 0.0 | 0.4 | 0.0 | 5.4 |
| | 2 | 0.0 | 0.0 | 0.1 | 0.0 | 2.1 |
| | 3 | 0.0 | 0.0 | 1.4 | 0.0 | 8.3 |
| | 4 | 0.5 | 0.0 | 0.0 | 0.0 | 0.3 |
| wild type vs hNR | 1 | 1.1 | 8.9 | 4.8 | 0.4 | 29.0 |
| | 2 | 0.0 | 2.6 | 6.6 | 1.2 | 29.5 |

The phenotype classifier classifies each connected region obtained from worm detector into Nile Red phenotypes. The connected regions do not always correspond to individual worms, as worms that touch or overlap would be in the same region. Yet we found that the connected regions are good units to classify into phenotypes in order to find the dominant phenotype of worms in a well. To predict the dominant phenotype of worms in a well, we classify all the connected regions in the well using the phenotype classifier and sum the scores that they regions got from the phenotype classifier. We use the sign of the total score to predict the dominant phenotype present in the well. The features used by the phenotype classifier are based on the size, shape and geometry of connected region and on the properties of the fluorescent stripes inside the region.

We tested our methods on images from plates with worms whose phenotype is known. We evaluate the performance on two binary classification tasks. One task is discriminating wells with wild type worms from lNR worms, and the other is discriminating wells with wild type worms from hNR worms. We report the average classification error rate for each plate using 2-fold cross validation.

In addition, we tested a classification method that abstains from predicting when the classifier has low confidence in its prediction. Our confidence rated classification method uses a bootstrap approach similar to Bagging (Breiman 1996). Instead of constructing a single master classifier, we construct a bagged master classifier by combining 7 different classifiers. Each of the 7 different classifiers is trained using a different random subset of the training examples. The bagged master classifier classifies a test instance in the following way. The instance is presented to each of the 7 classifiers. If all the 7 classifications are same, the master classifier outputs that classification. If there is a disagreement between the 7 classifications, then the master classifier abstains and does not predict. This decreases the classification error rate at the cost of having no prediction for some of the wells. This compromise is reasonable for HTS designs in which the cost associated with repeating an experiment is significantly lower than the cost associated with a screening mistake.

## 3 RESULTS

To evaluate our automated phenotype classifier, we compared our classifier performance to the performance of trained biologists in distinguishing lNR or hNR from wild type worms. We conducted six control experiments, each on a different 96-well plate. For each plate, half of the wells had wild type worms and the other half had a particular gene inactivated to give an lNR or hNR phenotype. For four of the plates, the gene inactivation made the worms exhibit the lNR phenotype, while for the two other plates, the gene inactivation increased the Nile Red signal in the worms.

We evaluate the performance of the phenotype classifier using a 2-fold cross validation (CV) method. Each CV corresponds to a random partition of the wells in a particular 96 well plate into two parts. Each part consists of 24 images of worms treated with hNR or lNR RNAi and 24 images of wild type worms. Every CV set is used for two evaluations, each using one part as the training set and the other part as the test set. The accuracy of our prediction for each plate is reported in Table 2, in the column titled "Automated method's error - without abstention". We compare the performance of our automated phenotype classifier to two biologists. Similar to the classifier evaluation, the biologists are shown half the images as training images and asked to classify the other set of images. The biologists were asked to classify the test images four times for each plate with a different random partition of the images.

The accuracy of our phenotype separator is similar to the accuracy of the biologists. The task of separating wild type from lNR is considerably easier than that of separating wild type from hNR. In both cases, the number of mistakes made by our system is similar to that of the worse of the two experimentalists.

In addition, we computed the system's error rate when allowing it to abstain from classifying some of the wells with low confidence classification (as defined above). The entries in the column titled "Automated method's error – with abstention" are the average number of mistakes when classifying only a "high-confidence" subset of images that exclude "ambiguous" or low-confidence images. The system labels an image "ambiguous" when there is disagreement among the component classifiers of the bagged master classifier. The disagreement between the component classifiers of the bagged master classifier is likely to be due to imaging or experimental issues such as improper focus or presence of very few worms in the image. The entries in the rightmost column, "Examples on which we abstain", are the average number of examples on which the system abstained. We see that not predicting the phenotype on the "ambiguous" images significantly improves the system's accuracy and makes it competitive with, and in most cases, better than that of human experts. The cost of this improvement is that in some cases about 25% - 30% of the images are not classified.

Researchers carrying out a high-throughput screen can significantly decrease the fraction of wells that are abstained from analysis by replicating the experiments of the wells, which in most cases is reasonable given that for most screens, experiments are done in duplicate or triplicate. The remaining small fraction of experiments that are ambiguous even with repetition could be easily screened manually.

## 4 DISCUSSION

To the best of our knowledge, the method described here the first effective computer vision method for distinguishing worm bodies from background on agar. We also present a novel method to classify fluorescent images of worms. While the phenotype classifier needs to be recalibrated for each plate, the worm segmentation algorithm is very stable, it did not require any additional calibration for the results presented here. We expect that the segmentation algorithm will not require significant recalibration in future experiments as long as the appearance of the worms in BF images does not change drastically. The fluorescence detector is designed to be specific for the Nile Red marker. However, the principles behind the detector, particularly, using features unique to the fluorescent marker to develop a robust classifier, should be widely applicable to the detection of a wide variety of fluorescent signals within a worm.

## 5 CONCLUSIONS

We present progress towards HTS-compatible analysis of images of worms on agar. Our novel worm segmentation approach successfully identifies worms in complex images and our fluorescence detector and phenotype classifier accurately distinguishes fluorescence-based phenotypes. We are currently working on applying this worm segmentation method to perform automated, high-throughput RNAi screens for Nile Red phenotypes in order to identify genes regulating aging in *C. elegans*.